\documentclass{article}

\usepackage{arxiv}

\usepackage[utf8]{inputenc}     
\usepackage[T1]{fontenc}        
\usepackage[hyphens]{url}       
\usepackage{graphicx}
\usepackage{microtype}
\usepackage{natbib}
\usepackage{caption}
\usepackage{amsmath}
\usepackage{amssymb}
\usepackage{amsfonts}
\usepackage{amsthm}
\usepackage{mathtools}          
\usepackage{mathrsfs}           
\usepackage{dsfont}             
\usepackage{multirow}
\usepackage{booktabs}           
\usepackage{array}              
\usepackage{makecell}
\usepackage{adjustbox}          
\usepackage{tabularx}
\usepackage{colortbl}
\usepackage[space]{grffile}     
\usepackage{subcaption}         
\usepackage{float}              
\usepackage{xcolor}
\usepackage{hyperref}           
\usepackage{doi}
\usepackage{tcolorbox}
\tcbuselibrary{most,skins,breakable}

\let\cite\citep

\setcitestyle{aysep={}}

\newcounter{eqboxcounter}

\newcommand{\eqboxlabel}[1]{\refstepcounter{eqboxcounter}\label{#1}}

\newcounter{textboxcounter}

\newtcolorbox{eqbox}[1][]{
    enhanced,
    breakable,
    colback=gray!5,
    colframe=black!70,
    boxrule=0.6pt,
    arc=3pt,
    left=10pt,
    right=10pt,
    top=8pt,
    bottom=8pt,
    #1
}

\newtcolorbox{summarybox}[1][]{
    enhanced,
    colback=blue!4,
    colframe=blue!55!black,
    boxrule=0.6pt,
    arc=4pt,
    left=10pt,
    right=10pt,
    top=8pt,
    bottom=8pt,
    fontupper=\small,
    #1
}

\definecolor{headerblue}{HTML}{1F3A5F}
\definecolor{rowtint}{HTML}{EEF3F8}
\definecolor{blocklabel}{HTML}{4A6FA5}

\newcolumntype{L}[1]{>{\raggedright\arraybackslash}p{#1}}

\title{Specifying Reward Functions for RL Without Environment Sampling}

\author{
	Stephane Hatgis-Kessell$^{1}$\thanks{Correspondence to: \texttt{stephhk@stanford.edu}} \quad
	W.~Bradley Knox$^{2}$ \quad
	Emma Brunskill$^{1}$ \\
	\normalfont\small $^{1}$Computer Science Department, Stanford University \\
	\normalfont\small $^{2}$Computer Science Department, The University of Texas at Austin
}

\renewcommand{\undertitle}{}
\renewcommand{\shorttitle}{Specifying Reward Functions for RL Without Environment Sampling}
\definecolor{linknavy}{HTML}{16336B}
\hypersetup{
colorlinks=true,
linkcolor=linknavy,
citecolor=linknavy,
urlcolor=linknavy,
pdftitle={Specifying Reward Functions for RL Without Environment Sampling},
pdfauthor={Stephane Hatgis-Kessell, W. Bradley Knox, Emma Brunskill},
}


\begin{document}

\maketitle  

\begin{abstract}
Enabling human stakeholders to specify reward functions that lead to their desired outcomes is a key challenge in deploying reinforcement learning agents. Preference-based methods such as online RLHF can reduce the burden of manual reward design, but they require repeatedly training policies, sampling trajectories from the real world, and eliciting feedback, making them impractical in settings where environment interaction is computationally expensive or unsafe. We introduce Experience-Free Autonomous Reward Specification (EARS), a method for learning reward functions from preferences without environment interaction. Our approach uses a structured LLM-mediated process to construct a small set of expressive reward features from a task description and the environment observation space, then strategically samples imagined trajectories in this feature space and learns feature weights from preferences over the imagined trajectory pairs. We evaluate on three long-horizon domains: pandemic lockdown regulation design, insulin administration for diabetes patients, and autonomous vehicle control on a highway. We compare EARS to baselines that also enable reward specification without environment interaction—namely, methods that directly prompt an LLM to generate a reward function. When learning from either ground-truth preference labels or preferences labeled by a LLM, EARS designs reward functions that are more aligned with the ground truth reward function that produced the preferences or LLM context than these baselines. These results suggest that preference-based reward specification remains effective without environment sampling, enabling practical reward design in settings where collecting real trajectories is costly or infeasible.
\end{abstract}

\section{Introduction}
\label{sec:intro}
A central challenge in AI is enabling human stakeholders to specify their objectives \citep{amodei2016concrete, krakovna2020specification,pan2022effects}. A common approach is to encode these objectives as a reward function and optimize it via reinforcement learning (RL). Unfortunately, hand-designing a reward function is difficult, and can often result in a decision policy that does not behave as intended \cite{amodei2016concrete,pan2022effects, krakovna2020specification}. This has motivated extensive interest in  learning a reward function from preferences such as in RLHF \citep{christiano2017deep, lee2021pebble, ibarz2018reward, pacchiano2021dueling, dong2024rlhf,sadigh2017active}. In particular, in \textit{online RLHF}, a reward model is learned from human preference data and used to train a policy (e.g., \cite{christiano2017deep}). The resulting policy is then rolled out in a real environment to generate trajectories. For example, the policy might control a humanoid learning to cook food, or generate a disaster response for extreme weather events modeled in a complex physics-based simulator. The resulting trajectories (of cooking, or disaster response in an expensive simulator) can then be used to elicit additional human feedback. The reward model is updated with this new data, and the process iterates. 

However, in many domains this process involves considerable cost. For example, for humanoid robotics environments or those that use expensive physics simulators, policy rollouts are expensive or risky. This can make both policy training given a fixed reward function and data collection of trajectories to elicit preferences over extremely costly. Therefore, when training a reward model requires many rounds of data collection, online RLHF can be difficult to apply in practice.

These challenges may also directly translate into implications for data-efficiency; long policy training times may require eliciting larger batches of preferences to limit the number of policy updates, which in turn may require more human preference labels to learn desirable behavior \citep{citovsky2021batch, biyik2024batch}. Further, constructing a dataset for preference elicitation requires sampling sufficiently diverse trajectories to elicit feedback over, including sub-optimal data. For many real-world domains, collecting large amounts of sub-optimal data is prohibitive due to safety concerns \citep{garcia2012safe}. If a high-fidelity simulator or world model is not available for data collection---which often require collecting large amounts of real world data to build---then acquiring the diverse and sufficiently sub-optimal trajectories required for effective preference elicitation may be very expensive \citep{cosner2022safety, liu2023efficient}. Offline-RLHF suffers from the same limitations, i.e., it requires sampling large amounts of sufficiently diverse experience. 

In this paper, we consider how to learn aligned reward functions without environment interaction—relevant when generating policy rollouts in the real domain is expensive. We introduce Experience-free Autonomous Reward Specification (EARS), which still assumes access to preference labels but learns a reward function without sampling from the environment. EARS comprises two novel stages. The first stage is a structured interaction between Large Language Models (LLMs) to construct a set of reward features that can represent different reward functions, following a process human domain experts might follow. Then in the second stage we strategically sample imagined trajectories from this generated feature space to elicit preferences over and construct a reward function. 

We evaluate EARS in three complex, long horizon decision-making tasks with preferences labeled either synthetically by a ground-truth reward function, or by an LLM conditioned on a natural language specification of the task objectives—the kind of document a careful human stakeholder could plausibly produce. We first show that prompting an LLM to generate a reward function—even with an exceedingly specific description of the ground-truth reward—often fails to produce a well aligned reward function, highlighting the limitations of the only alternative available when environment interaction is unavailable. We then show that EARS produces reward functions far more closely aligned with the ground truth when learning from ground-truth preferences or LLM labeled preferences. Moreover, EARS matches or outperforms methods that do sample real trajectories from the environment to elicit preferences over.
 
Our contribution is three-fold:
\begin{itemize}
    \item We show that directly prompting an LLM to design a reward function with an extremely detailed natural language description often does not produce a more aligned reward function than providing no reward function description. 
    \item We introduce EARS, to design reward functions with preferences without sampling from an environment. EARS samples zero environment transitions, making it strictly more environment-sample-efficient than methods like online RLHF. We show that when learning from ground-truth preferences it matches the preference-label efficiency of a baseline that learns from real trajectories and outperforms directly prompting an LLM.
    \item We demonstrate that, when only given access to a natural language specification of the task objectives, EARS designs more aligned reward functions than directly prompting an LLM with the same specification
\end{itemize}

We release all code to reproduce experiments in this work \href{https://github.com/Stephanehk/EARS}{here}. 


\section{Preliminaries}
\label{sec:prelims}
Consider an MDP $\mathcal{M} \triangleq (S, A, \Omega, \gamma, p_0, r)$ with state space $S$, action space $A$, transition dynamics $\Omega: S \times A \rightarrow \Delta(S)$, discount factor $\gamma \in [0,1]$, and initial state distribution $p_0$. Let $\tau$ denote a trajectory $\tau = (s^{\tau}_{0}, a^{\tau}_{0}, s^{\tau}_{1}, a^{\tau}_{1}, \ldots)$ starting at $s^{\tau}_{0} \sim p_0$. We allow $\tau$ to be finite (e.g., when terminating in an absorbing state) or infinite. In this work, we consider the undiscounted setting ($\gamma=1$); the formulation extends naturally to the discounted case ($\gamma<1$).
The ground-truth reward function is $r: S \times A \times S \rightarrow \mathbb{R}$. We denote by $\mathcal{M} \setminus r \triangleq (S, A, \Omega, \gamma, p_0, \_\,)$ the environment without a specified reward function. Let $r$ be the (unobservable) ground-truth reward function, $\hat r$ a learned approximation, and $\tilde r$ an arbitrary reward function. We denote the reward function produced by EARS as a linear functions over features: $\hat r(s,a,s') = \hat w^\top \phi(s,a,s')$
where $\phi(s,a,s') \in \mathbb{R}^d$ is a feature representation and $\hat w \in \mathbb{R}^d$ is a weight vector. We assume EARS produces a linear reward function to enable its active learning procedures in outlined Section \ref{sec:ears_stage2}. Importantly, this assumption does not limit our evaluations of EARS; the ground-truth reward functions and the reward function produced by the other methods we consider may be non-linear. We do not assume access to a known feature representation that can be used to construct the ground-truth reward function; this work concerns both learning $\hat{w}$ and specifying $\phi$.
A policy $\pi: S \times A \rightarrow [0,1]$ maps states to action distributions. Its expected discounted return under $\tilde r$ from start distribution $p_0$ is $J_{\tilde r}(\pi)$. An \emph{optimal policy} for $\tilde r$ is any $\pi^*_{\tilde r} \in \arg\max_\pi J_{\tilde r}(\pi)$.  

As a complement to $\tau$, let $\Phi$ denote a trajectory expressed as the sum of discounted reward features. For example, for a given trajectory, $\Phi_\tau = \sum_{t=0}^{|\tau|-1} \gamma^t \phi(s_t^\tau, a_t^\tau, s_{t+1}^\tau)$ where $|\tau|$ denotes the number of transitions in $\tau$ (possibly infinite). Going forward, we will only consider trajectories expressed as the sum of reward features, rather than as a list of transitions. We learn a reward function $\hat r$ from trajectory pair preferences. Let $\mathcal{D}=\{(\Phi_1,\Phi_2,\mu)\}_{k=1}^N, \quad \text{and }
\mu \in \{0,1,\tfrac12\} \text{ with }
0:\Phi_1\succ\Phi_2,\;
1:\Phi_2\succ\Phi_1,\;
\tfrac12:\Phi_1\sim\Phi_2.$ where $\succ$ means is ''preferred to" and $\sim$ means is ''equally preferred to".

We consider two different approaches to learning a reward function from the preference dataset. 

\textbf{Learning from the Bradley–Terry preferences~~} A standard assumption in RLHF is that preferences are generated by the  Bradley–Terry preference model
\citep{christiano2017deep}:
\begin{equation}
\begin{gathered}
   P(\Phi_1\succ \Phi_2 \mid \tilde w)=\sigma\!\big(\tilde w^T\Phi_1-\tilde w^T\Phi_2\big),\\
\sigma(x)=1/(1+e^{-x})
\end{gathered}
\label{eq:btl_df}
\end{equation}

and that $\hat w$ is learned by minimizing the cross-entropy loss:

\begin{equation}
\label{eq:loss}
\begin{aligned}
\mathcal{L}_{\text{pref}}(\hat{w};\mathcal{D}_t) =
- \hspace{-3.5mm}\sum_{(\Phi_1, \Phi_2, \mu) \in \mathcal{D}} \hspace{-2mm}\big[ & \mu \log {P}(\Phi_1 \succ \Phi_2 | \hat{w}) \\
& + (1-\mu) \log {P}(\Phi_1 \prec \Phi_2 | \hat{w}) \big].
\end{aligned}
\end{equation}

While widely used, the assumption that preferences are generated according to the Bradley--Terry model can be limiting; see \cite{zhixuan2024beyond} for a detailed critique. Nevertheless, when evaluating reward specification from ground-truth preferences---i.e., preferences generated by a ground-truth reward function to approximate those a real human might provide---we adopt the Bradley--Terry model and learn a reward function by minimizing the cross-entropy loss.

\textbf{Learning from noiseless preferences~~} Assuming preferences are sampled from the Bradley-Terry preference model relies on the assumption that preferences follow a particular type of structured noise. Here, we outline a method for learning a reward function from preferences that are noiselessly generated by a reward function, as has been assumed by some prior work \citep{knox2206models, hatgis2501influencing}. In particular, we assume:
\begin{equation}
    P(\Phi_1 \succ \Phi_2 \mid \tilde w)
=
\mathds{1}\!\left\{\tilde w^T \Phi_1 > \tilde w^T \Phi_2\right\},
\label{eq:det_pref_model_def}
\end{equation}
When preferences are labeled by an LLM, rather than by humans or synthetic annotators, we posit that this constitutes a strong alternative to the Bradley--Terry assumptions, which were originally developed to model human behavior---albeit are still flawed \citep{zhixuan2024beyond}. Following the observation of \cite{kim2024unified}, given preference dataset $\mathcal{D}$, we can learn a reward function by solving the following linear program:
\begin{equation}
\label{eq:reward_lp}
\begin{aligned}
\text{find} \quad & \hat{w} \in \mathbb{R}^d \\
\text{s.t.} \quad
& \hat{w}^\top \left(\Phi_{1,k} - \Phi_{2,k}\right) \geq \epsilon,
&& \forall k \text{ where } \mu_k = 0, \\
& \hat{w}^\top \left(\Phi_{1,k} - \Phi_{2,k}\right) \leq \epsilon,
&& \forall k \text{ where } \mu_k = 1, \\
& \hat{w}^\top \left(\Phi_{1,k} - \Phi_{2,k}\right) = 0,
&& \forall k \text{ where } \mu_k = \tfrac12, \\
& \|\hat{w}\|_1 \leq B.
\end{aligned}
\end{equation}

where $\epsilon$ is a tolerance parameter. 

\section{Experience Free Automatic Reward Specification (EARS)}
\begin{figure}[t]
    \centering
    \begin{minipage}[t]{0.53\textwidth}
    \vspace{0pt}
    \includegraphics[width=\textwidth]{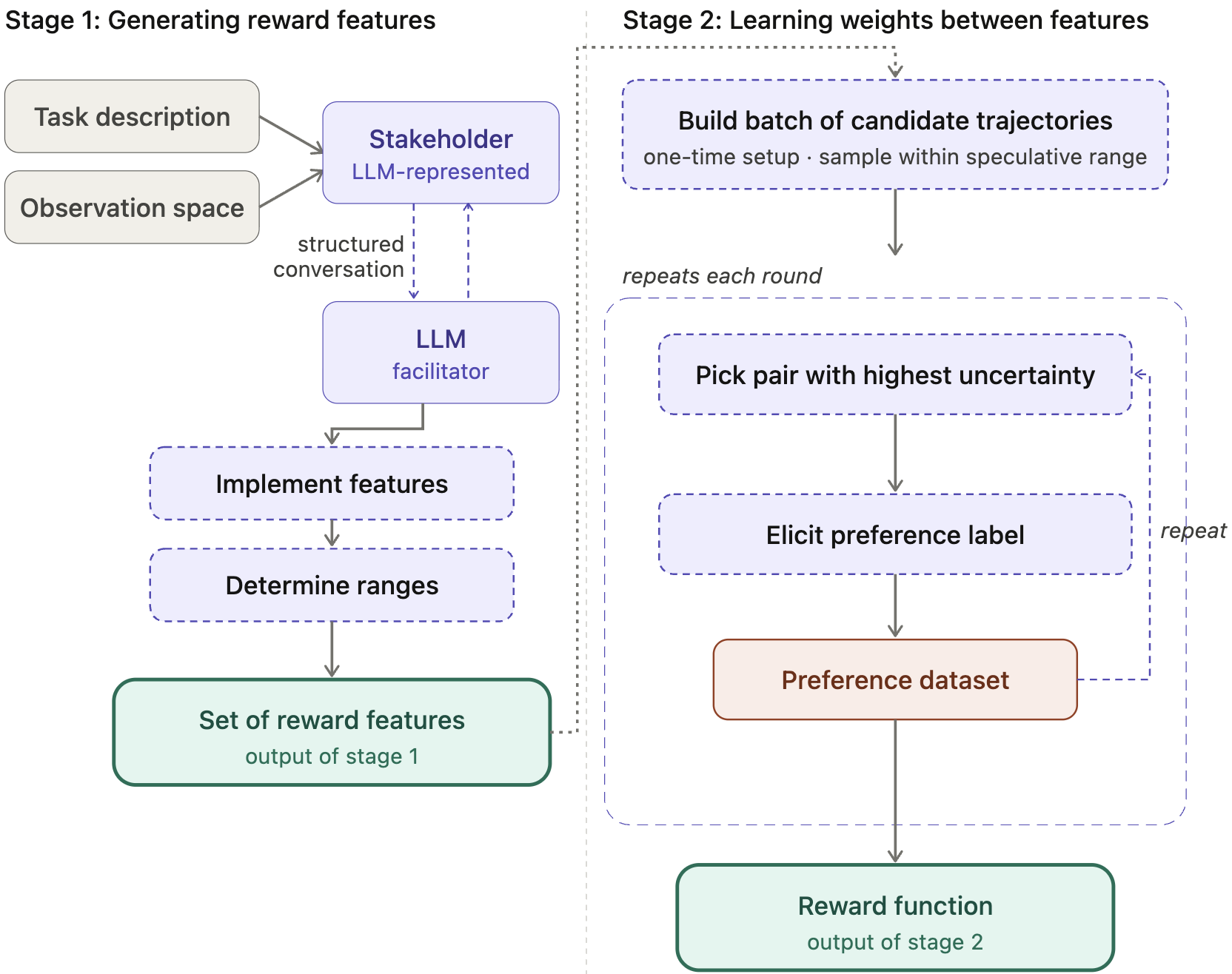}
    \end{minipage}\hfill
    \begin{minipage}[t]{0.45\textwidth}
     \vspace{0pt}
        \caption{\textbf{Experience-free Autonomous Reward Specification (EARS)}: In Stage 1, a natural language task description and a description of the environment observation space are inputted to an LLM-represented stakeholder. The LLM-represented stakeholder deliberates with an LLM-represented facilitator to design a small set of reward features that can be combined linearly to represent different objectives. In Stage 2, imagined trajectory pairs are sampled from the designed feature space and preferences are labeled over those pairs. The output of Stage 2 is a reward function learned from those preferences.}
        \label{fig:method_overview}
    \end{minipage}
\end{figure}
EARS consists of two stages, illustrated in Figure \ref{fig:method_overview}. In the first stage, we engage an LLM, representing a human stakeholder or group of stakeholders, in a structured conversation with another LLM, representing a facilitator. The aim of the conversation is to design a small set of expressive and interpretable features that, when combined linearly, can represent the different objectives a stakeholder may have. In the second stage we elicit preferences to learn the weights between those reward features. 
%
The output of the second stage is a reward function learned to satisfy the elicited preferences. The environment is never sampled to produce the reward function.

Consider designing a reward function to train a policy to determine lockdown regulations in a Covid-19 pandemic. In Stage 1, the stakeholder and facilitator LLMs deliberate over a task description ("set COVID-19 lockdown regulations") and the observation space to produce a small set of measurable reward features—say, the fraction of the population in a critical condition, the severity of the current regulation stage, and the hospital capacity---each outputted by a Python function with a speculative range. In Stage 2, we sample imagined trajectories in this feature space (e.g., "what if'' questions such as those comparing a low critical-case count under sustained strict lockdown versus a higher count under less stringent regulations), elicit preferences over such pairs, and learn the weights that trade these features off, yielding a reward function without rolling out any policies. 

\subsection{Stage 1: Designing a set of reward features}

The input to stage 1 is a brief description of the task, a description of the environment observation space represented as Python code, and optionally, a description of the task objective, supplied by a human stakeholder. The output of stage 1 is a small set of reward features: $\phi(s_t,a_t,s_{t+1}) = \{\phi_1(s_t,a_t,s_{t+1}), ..., \phi_d(s_t,a_t,s_{t+1})\}$. Each reward feature $\phi_i$ is implemented as a python function, and is later linearly combined to produce a reward function. 

To generate the set of reward features, we engage a stakeholder and facilitator, both represented as LLMs, in a structured conversation. The conversation seeks to: (1) unearth the different objectives people may have, (2) determine which objectives are measurable given the environment's observation space, and (3) aggregate and implement the identified objectives as executable Python functions. Additionally, to enable the sampling procedure used in stage 2, stage 1 also outputs a speculative minimum and maximum range
of the sum of each reward feature over of a trajectory of $H$ steps---denoted as $\phi_{\text{min}}^H \in \mathbb{R}^d$ and $\phi_{\text{max}}^H \in \mathbb{R}^d$. We prompt the stakeholder to select finite values that represent the likely ranges that would be observed. 

\subsection{Stage 2: Learning weights over the reward features}
\label{sec:ears_stage2}
The input to stage 2 is the set of generated reward features from stage 1 and a preference labeler---whether that be a human stakeholder, an LLM, or something else. The output of stage 2 is a set of weights over the set of generated reward features, i.e., a reward function. 

\textbf{Constructing imagined trajectories~~} Our key insight is that, given a small set of generated reward features, we can construct “imagined” trajectories by sampling feature values uniformly within their feasible ranges. Specifically, for each feature, we uniformly sample values from a speculated minimum–maximum range determined in Stage 1. We denote an imagined trajectory as $\Phi \in \mathbb{R}^d$, where $d$ is the number of reward features, which is constructed by sampling from the uniform distribution $\mathcal{U}$ over reward feature ranges: $\Phi \sim \mathcal{U}([\phi_{\min}^H, \phi_{\max}^H])$. While many of these imagined trajectories may not be physically plausible, preferences over them can still be useful for learning a reward function.

\textbf{Strategically sampling imagined trajectories~~} To avoid requiring prohibitively large datasets of preferences, prior work in online RLHF focuses on strategically sampling trajectory pairs that can reduce uncertainty \citep{christiano2017deep, lee2021pebble, ibarz2018reward, pacchiano2021dueling, dong2024rlhf,sadigh2017active}. This approach is particularly apt for imagined trajectories, since sampling them is cheap and we can draw many candidates and select the single most informative pair. At each round of feedback collection, we sample $k$ imagined trajectories and then select the pair with the highest uncertainty under the current dataset $\mathcal{D}$, according to an uncertainty function $f(\Phi_1, \Phi_2, \mathcal{D})$: $(\Phi_1^*, \Phi_2^*) = \arg\max_{\Phi_1, \Phi_2} f(\Phi_1, \Phi_2, \mathcal{D})$ where $f(\Phi_1, \Phi_2, \mathcal{D})$ is the subsequent sections. We restrict the selection to pairs of imagined trajectories that differ in at most two features; we speculate that this makes the salient differences between any two imagined trajectories easier for a preference labeler to reason about. A preference label is elicited over the selected imagined trajectory pair, added to the preference dataset, and then the process is repeated. $f(\Phi_1, \Phi_2, \mathcal{D})$ can be instantiated using any uncertainty measure proposed in prior work on online-RLHF. In this work, we consider two such measures, depending on whether preferences are assumed to be stochastic or noiseless. 

\textbf{Strategically sampling imagined trajectories with stochastic preferences~~}
When assuming preferences follow the Bradley-Terry preference model, we quantify uncertainty as the variance in predicted preference probabilities across an ensemble of reward models. In this work, the ensemble is maintained using an Epistemic Neural Network (ENN)~\citep{osband2023epistemic} which provides a computationally efficient method for measuring epistemic uncertainty as the variance across different model predictions. Concretely, let $\{\hat{r}_{\theta^{(i)}}\}_{i=1}^M$ denote the ensemble of reward models, and define
\[
p^{(i)}(\Phi_1 \succ \Phi_2)
= \frac{\exp\big(\hat{r}_{\theta^{(i)}}(\Phi_1)\big)}{\exp\big(\hat{r}_{\theta^{(i)}}(\Phi_1)\big) + \exp\big(\hat{r}_{\theta^{(i)}}(\Phi_2)\big)}.
\]
Then, $f_{\text{enn}}(\Phi_1, \Phi_2, \mathcal{D})
= \mathrm{Var}_{i \in [M]}\left[ p^{(i)}(\Phi_1 \succ \Phi_2) \right]$. This variance-based uncertainty measure is common in prior work \citep{christiano2017deep, lee2021pebble, dwaracherla2024efficient}.

\textbf{Strategically Sampling Imagined Trajectories with Noiseless Preferences~~} Alternatively, when assuming preferences are noiseless and learned by solving a linear program via Eq. \ref{eq:reward_lp}, we quantify uncertainty using the disagreement among the reward functions that are still consistent with $\mathcal{D}$. Each candidate reward function $w$ assigns the pair a signed return difference $w^\top (\Phi_1 - \Phi_2)$, whose sign encodes which trajectory $w$ prefers. Let $\mathcal{W}(\mathcal{D})$ denote the feasible set of weights satisfying the constraints from $\mathcal{D}$, i.e., the solutions to Eq. \ref{eq:reward_lp}, and let $\ell$ and $u$ be the smallest and largest such differences over this set:
\[
\ell = \min_{w \in \mathcal{W}(\mathcal{D})} w^\top (\Phi_1 - \Phi_2),
\quad
u = \max_{w \in \mathcal{W}(\mathcal{D})} w^\top (\Phi_1 - \Phi_2).
\]
We then define
\[
f_{\mathrm{LP}}(\Phi_1, \Phi_2, \mathcal{D})
=
\begin{cases}
u - \ell & \text{if } \mathrm{sign}(\ell) \neq \mathrm{sign}(u), \\
0 & \text{otherwise.}
\end{cases}
\]
A differing sign means some feasible reward function prefers $\Phi_1$ while another prefers $\Phi_2$, so the pair is contested and $f_{\mathrm{LP}}$ measures the magnitude of that disagreement. When $\mathrm{sign}(\ell) = \mathrm{sign}(u)$ (e.g., $\ell = 1$, $u = 10000$), every feasible reward function assigns the difference the same sign and therefore agrees on the ordering; the pair is uninformative and we set its uncertainty to zero.\footnote{As a fallback, if no trajectory pairs are found with non-zero uncertainty, we do not zero out the uncertainty.}

\textbf{Learning a reward function from preferences over imagined trajectories~~} The resulting dataset $\mathcal{D}$ consists of tuples $(\Phi_1, \Phi_2, \succ)$. A reward function is learned from this dataset using either the objective in Eq.~\ref{eq:loss} when assuming Bradley--Terry preferences, or the linear program in Eq.~\ref{eq:reward_lp} when assuming noiseless preferences. Let $\hat{w}_{\mathcal{D}}$ be the weight vector that satisfies the preferences between imagined trajectory pairs in dataset $\mathcal{D}$ found with either reward learning method.\footnote{When solving the linear program in Eq. \ref{eq:reward_lp}, $\hat{w}_{\mathcal{D}}$ need not be unique.} The transition-level reward function, which inputs transitions from the environment, is then $\hat r(s_t, a_t, s_{t+1})
= \hat{w}_{\mathcal{D}}^\top \phi(s_t, a_t, s_{t+1})$.
%



\section{Experiments}
We now empirically evaluate EARS. We consider two ways a stakeholder may specify a reward function: via preferences (Section~\ref{sec:synth_pref_rew_spec})---which we study by simulating human preferences with a ground-truth reward function and two different preference models---and via a natural language specification of the task objectives, which serves as context for an LLM-based preference labeler (Section~\ref{sec:reward_desc_rew_spec}). This second setting represents a directly deployable use of EARS: the only inputs required are an environment description and a specification document a careful human could write, with no environment sampling and no human-in-the-loop preference labeling. All experiments use Gemini~3.

\subsection{Evaluating the learned reward function.}
\label{sec:eval_learned_rew}
We evaluate the alignment between a learned reward function $\hat{r}$ and the ground-truth reward function $r$ using the Trajectory Alignment Coefficient (TAC) \citep{muslimani2025towards}. TAC computes the Kendall-Tau correlation between the ranking of trajectories under the learned reward function,$\hat{r}$, and the ground truth reward function, $r$, where all trajectories share the same start state. We compute TAC using $500$ trajectories sampled from the environment, denoted $\mathcal{T}_{\text{eval}}$. These trajectories are collected from checkpoints while training a policy with the ground-truth reward function, and are used solely for evaluation; consistent with our problem setting, no environment samples are used for learning the reward function. Our aim is to develop a method to achieve the following:

\begin{tcolorbox}[colback=gray!10,colframe=gray!50,boxrule=0.5pt,arc=2pt,
                  left=4pt,right=4pt,top=2pt,bottom=2pt]
\small
\textbf{Objective.} $\max_{\hat{r}} \; \mathrm{TAC}(\hat{r}, r)
\;\; \text{s.t.} \;\;
\hat{r}$ is learned without sampling transitions from $\mathcal{M}$.
\end{tcolorbox}

We further discuss this evaluation metric in relation to others in Section \ref{app:sec:eval_learned_rew}.

\subsection{Environments}

We evaluate EARS across three benchmark environments used in \citet{pan2022effects}, summarized in Table \ref{tab:envs}. For each environment, we consider two different ground-truth reward functions to align with, denoted as $r_1$ and $r_2$. These two ground-truth reward functions per environment are also designed by \citet{pan2022effects} and are given in Section \ref{app:gt_rew_desc}. All environments have large, continuous observation spaces that must be reasoned over to design a small set of reward features.

\begin{table}[t]
\centering
\small
\setlength{\tabcolsep}{2.5pt}
\renewcommand{\arraystretch}{0.95}
\begin{adjustbox}{max width=\textwidth}
\begin{tabular}{L{1.9cm} L{2.6cm} c L{1.4cm} L{1.7cm} L{3.2cm} L{3.4cm} L{2.3cm}}
\toprule
\textbf{Name} & \textbf{Objective} & \textbf{H} & \textbf{State} & \textbf{Action} &
\textbf{$r_1$ Summary} & \textbf{$r_2$ Summary} & \textbf{Simulator} \\
\midrule
\textbf{Pandemic Mitigation} \citep{kompella2020reinforcement} & 
Design COVID-19 pandemic lockdown regulations &
192 &
Cont., 312-dim &
Disc. $\{-1, 0, 1\}$ &
Trades off between lockdown regulation strictness and infection rates &
Encourages strict and non-fluctuating lockdown regulations even when infection rates are low &
Modified SEIR simulator \citep{kompella2020reinforcement} \\
\midrule
\textbf{Glucose Monitoring}\newline \citep{dallaman2014uvapadova,fox2020deep} & 
Administer insulin to patient with Type II diabetes &
5760 &
Cont., 96-dim &
1-d Cont. in $[0, 1]$ &
Prioritizes minimizing health risks &
Prioritizes reducing financial cost of treatment and risk of death &
FDA-approved simulator \citep{dallaman2014uvapadova,fox2020deep} \\
\midrule
\textbf{Traffic Control}\newline \citep{wu2021flow} & 
Control autonomous vehicle (AV) fleet on highway &
300 &
Cont., 50-dim &
10-d Cont. in $[0, 1]$ &
Trades off between maximizing mean velocity of all vehicles and maintaining safe distances &
Maximizes mean velocity &
FLOW highway simulator \citep{wu2021flow} \\
\bottomrule
\end{tabular}
\end{adjustbox}
\caption{Environments, used by \citet{pan2022effects}. H = horizon.}
\label{tab:envs}
\end{table}

\subsection{Baselines}
\label{sec:baselines}

\textbf{Baselines that don't sample from environment~~} To the best of our knowledge, the only existing class of methods that specifies a reward function without environment sampling is to directly prompt an LLM to write a reward function (e.g., \cite{ma2023eureka, kwon2023reward, xie2023text2reward, yu2023language}). These methods typically then refine the generated reward function through environment sampling—for example, by executing the proposed reward, observing the resulting behavior, and iterating. We introduce three baselines in this class but with the environment-sampling step removed to match our problem setting. Each baseline takes as input a task description and a description of the environment observation space, and outputs a reward function implemented in Python. Two of the baselines supply the reward-designing LLM with a natural language specification describing the task objectives and how they should be traded off---the same specifications that we evaluate EARs with in Section \ref{sec:reward_desc_rew_spec}. 


\begin{itemize}
    \item \textbf{Direct Prompting:} Prompts an LLM to design a reward function for an RL agent, given only the task description and observation space. No stakeholder specification is provided.
    
    \item \textbf{Direct Prompting + Realistic Reward Description:} The reward-designing LLM is given a realistic stakeholder specification, produced by a separate LLM that fills out a structured template using a summary of the ground-truth reward. This specification represents what we believe a careful human stakeholder could plausibly produce in the realistic setting where no ground-truth reward function exists. The realistic reward descriptions are given in Appendix Figures \ref{fig:pandemic_r1_pt2}, \ref{fig:glucose_r1_pt2}, and \ref{fig:traffic_r1_pt2}.
    
    \item \textbf{Direct Prompting + Privileged Reward Description:} The reward-designing LLM is given a highly detailed objective specification. To generate this specification, a separate LLM is shown the ground-truth reward implementation and asked to describe it in natural language. The resulting specification is likely more detailed than what a real human stakeholder without access to a ground-truth reward function could specify; it serves as an upper bound on the specificity available to a direct-prompting method. The privileged reward descriptions are given in Appendix Figures \ref{fig:pandemic_r1_pt1}, \ref{fig:glucose_r1_pt1}, and \ref{fig:traffic_r1_pt1}.
\end{itemize}

In real-world use, the specification for the last two baselines would come directly from a human stakeholder, and no ground-truth reward function would exist. 

\textbf{Baselines that do sample from the environment~~} We compare EARS to two baselines that, unlike EARS, learn from real trajectories sampled from the environment.  Both baselines learn from preferences labeled by the ground-truth reward function over the same real trajectory pairs. The trajectories are sampled uniformly from a candidate pool built by rolling out policy checkpoints saved while training a policy with the ground-truth reward function, and pairs are uniformly sampled from this set.
The two baselines differ only in the trajectory representation input to the model they learn; both learn a model parameterized as a non-linear neural network. Offline-RLHF learns a reward model defined over transitions, which can be used directly as a reward function. Learning To Rank Real Trajectories (LTRRT) instead represents each trajectory as the sum of its observation-space features, potentially avoiding challenges in correctly assigning trajectory return across transitions; because this model is defined over a full trajectory rather than transitions, it cannot itself be used as a reward function, and serves only to rank trajectories. \footnote{We can still compare LTRRT to EARS because the TAC, our primary evaluation metric, only inputs an ordering over trajectories}
Note that both baselines enjoy an advantage unavailable to EARS: their preference pairs are drawn from real trajectories produced by a policy trained on the ground-truth reward function, whereas EARS---with no access to the ground-truth reward function---only constructs imagined trajectories. More details are provided in Appendix \ref{app:offline-pref-learning}. 


\subsection{Results when specifying a reward function via ground-truth preferences}
\label{sec:synth_pref_rew_spec}
We evaluate EARS when assuming preferences are sampled from the Bradley-Terry preference model under the ground-truth reward function and when assuming preferences are noiselessly generated by the ground-truth reward function. Because the ground-truth reward function inputs real transitions from the environment, but we elicit preferences over imagined trajectories, we learn a mapping between the two representations as detailed in Appendix Section \ref{app:imagined_pref_mapping}. 

\begin{table}[t]
\centering
\caption{\textbf{EARS with preferences compared to prompting baselines:~~} TAC between learned and ground-truth reward functions across three environments, evaluated for two ground-truth reward functions ($r_1$ and $r_2$). We compare EARS against three prompting baselines: direct prompting with no reward description (No-RD), with a privileged reward description (Privileged-RD) and with a realistic reward description (Realistic-RD). We evaluate EARS with noiseless preferences and stochastic preferences. None of the methods sample transitions from the environment. Results are averaged over three random seeds ($\pm$ standard error), with each seed rerunning the full method including all LLM generations. Higher TAC indicates closer alignment with the ground-truth reward function.}
\label{tab:env_free_reward_results}
\setlength{\tabcolsep}{4pt}
\renewcommand{\arraystretch}{1.05}
\begin{tabular}{llcccccc}
\toprule
& & \multicolumn{2}{c}{Pandemic} & \multicolumn{2}{c}{Glucose} & \multicolumn{2}{c}{Traffic} \\
\cmidrule(lr){3-4} \cmidrule(lr){5-6} \cmidrule(lr){7-8}
Method & Variant & $r_1$ & $r_2$ & $r_1$ & $r_2$ & $r_1$ & $r_2$ \\
\midrule
\multirow{2}{*}{EARS} & 1k noiseless prefs. & $\boldsymbol{.72\!\pm\!.01}$ & $\boldsymbol{.96\!\pm\!.00}$ & $\boldsymbol{.69\!\pm\!.07}$ & $\boldsymbol{.82\!\pm\!.04}$ & $\boldsymbol{.92\!\pm\!.07}$ & $\boldsymbol{.90\!\pm\!.09}$ \\
 & 2k stochastic prefs. & $\boldsymbol{.73\!\pm\!.01}$ & $\boldsymbol{.94\!\pm\!.02}$ & $\boldsymbol{.74\!\pm\!.03}$ & $.71\!\pm\!.06$ & $\boldsymbol{.85\!\pm\!.04}$ & $\boldsymbol{.90\!\pm\!.09}$ \\
\midrule
\multirow{3}{*}{Direct Prompting} & No-RD & $-.08\!\pm\!.12$ & $.43\!\pm\!.12$ & $.16\!\pm\!.20$ & $-.20\!\pm\!.20$ & $.25\!\pm\!.06$ & $.67\!\pm\!.27$ \\
& Realistic-RD & $-.37\!\pm\!.10$ & $.68\!\pm\!.06$ & $.27\!\pm\!.29$ & $\boldsymbol{.79\!\pm\!.14}$ & $.27\!\pm\!.04$ & $\boldsymbol{.94\!\pm\!.01}$ \\
 & Privileged-RD & $-.14\!\pm\!.14$ & $.58\!\pm\!.19$ & $.21\!\pm\!.17$ & $\boldsymbol{.84\!\pm\!.12}$ & $.17\!\pm\!.01$ & $\boldsymbol{1.00\!\pm\!.00}$ \\
\bottomrule
\end{tabular}
\end{table}

Results are shown in Table~\ref{tab:env_free_reward_results}, demonstrating that EARS learns more aligned reward functions from the ground-truth preferences than other methods that also do not rely on environment sampling.
Notably, even providing the Direct Prompting baseline with an exceedingly specific objective description for $r_1$ (Privileged-RD) fails to generate an aligned reward function in all environments. EARS slightly under performs the direct prompting baselines when all methods produce reward functions with higher TAC, and substantially outperforms them otherwise. 
\begin{table}[t]
\centering
\caption{\textbf{EARS with natural language input compared to prompting baselines:~~}We evaluate EARS with preferences labeled by an LLM given a natural language specification generated by the ground-truth reward function against direct prompting baselines that generate reward functions from the same specification. We evaluate with two specifications outlined in Section \ref{sec:baselines}: (a) Realistic Reward Description and (b) Privileged Reward Description. See Table \ref{tab:env_free_reward_results} for more details.}
\label{tab:env_free_reward_results_llm_prefs}
\setlength{\tabcolsep}{7pt}
\renewcommand{\arraystretch}{0.95}
\small
\begin{tabular}{llcccccc}
\toprule
& & \multicolumn{2}{c}{Pandemic} & \multicolumn{2}{c}{Glucose} & \multicolumn{2}{c}{Traffic} \\
\cmidrule(lr){3-4} \cmidrule(lr){5-6} \cmidrule(lr){7-8}
Specification & Variant & $r_1$ & $r_2$ & $r_1$ & $r_2$ & $r_1$ & $r_2$ \\
\midrule
\multirow{2}{*}{(a) Realistic} & Direct Prompting & $-.37\!\pm\!.10$ & $.68\!\pm\!.06$ & $.27\!\pm\!.29$ & $.79\!\pm\!.14$ & $\boldsymbol{.27\!\pm\!.04}$ & $\boldsymbol{.94\!\pm\!.01}$ \\
 & EARS & $\boldsymbol{.43\!\pm\!.16}$ & $\boldsymbol{.74\!\pm\!.02}$ & $\boldsymbol{.85\!\pm\!.10}$ & $\boldsymbol{.89\!\pm\!.04}$ & $-.03\!\pm\!.14$ & $\boldsymbol{.95\!\pm\!.02}$ \\
\midrule
\multirow{2}{*}{(b) Privileged} & Direct Prompting & $-.14\!\pm\!.14$ & $.58\!\pm\!.19$ & $.21\!\pm\!.17$ & $\boldsymbol{.84\!\pm\!.12}$ & $.17\!\pm\!.01$ & $\boldsymbol{1.00\!\pm\!.00}$ \\
 & EARS & $\boldsymbol{.32\!\pm\!.08}$ & $\boldsymbol{.71\!\pm\!.07}$ & $\boldsymbol{.84\!\pm\!.00}$ & $\boldsymbol{.95\!\pm\!.00}$ & $\boldsymbol{.20\!\pm\!.20}$ & $.97\!\pm\!.03$ \\
\bottomrule
\end{tabular}
\end{table}

\begin{figure}[t]
    \centering
    \includegraphics[width=0.8\linewidth]{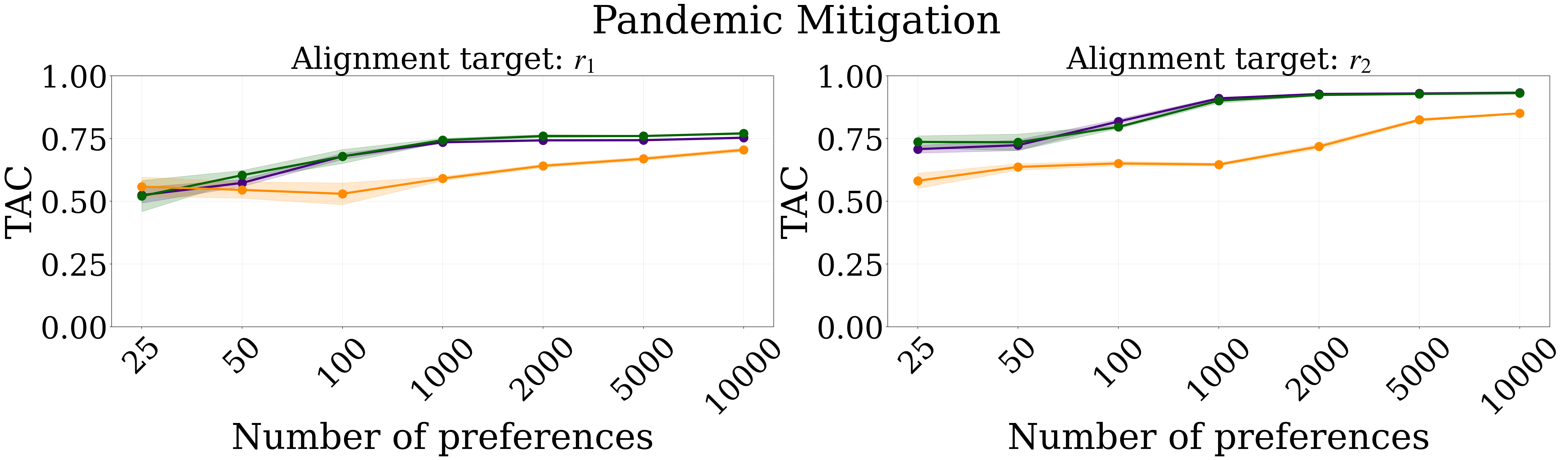}\\[0.4em]
    \includegraphics[width=0.8\linewidth]{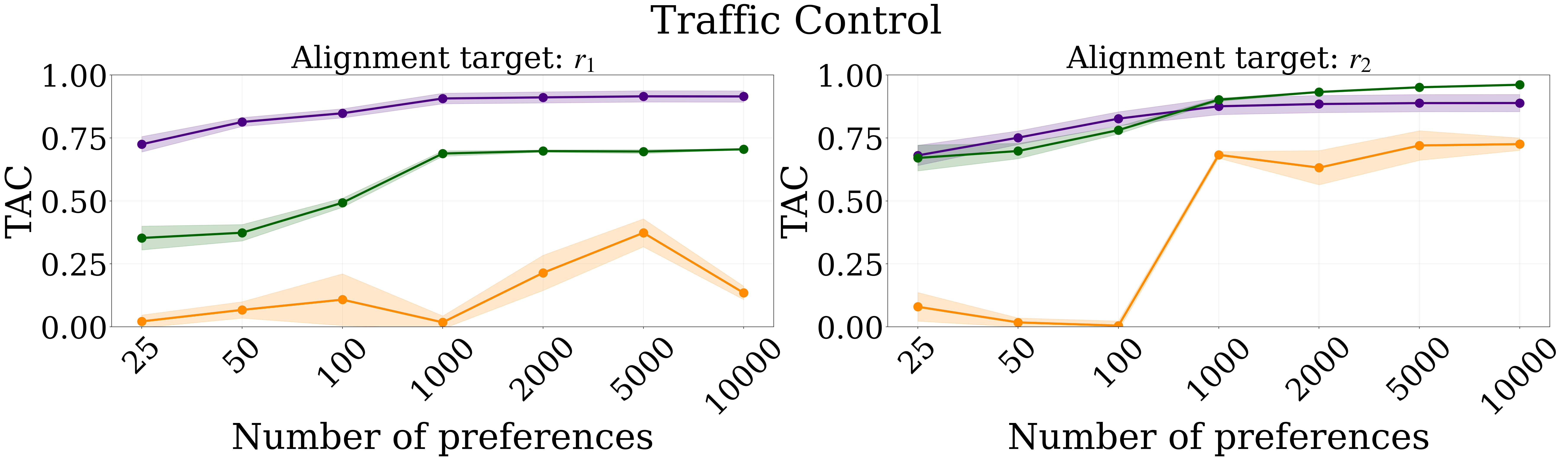}\\[0.4em]
    \includegraphics[width=0.8\linewidth]{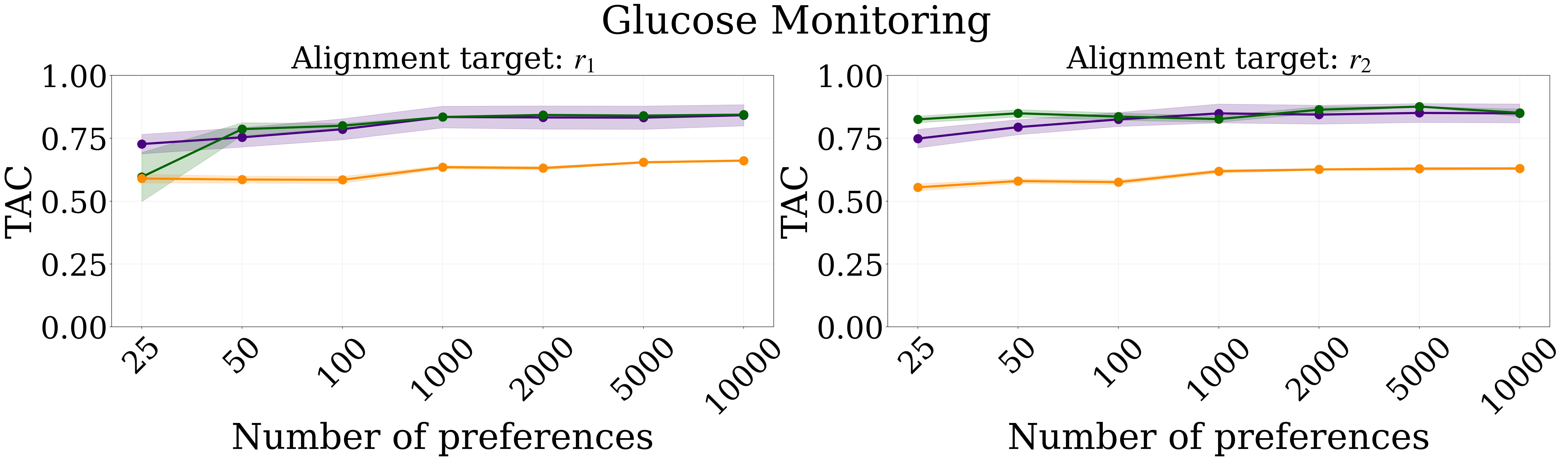}\\[0.5em]
    \includegraphics[width=0.6\linewidth]{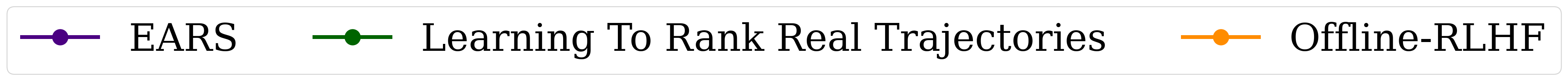}
        \caption{\textbf{Comparison to learning with real trajectories:~~} We compare EARS, which learns from preferences over imagined trajectories, against Offline-RLHF and Learning To Rank Real Trajectories, which both learn a non-linear model from trajectory pairs sampled from policy checkpoints in the environment. Results are shown for 5 seeds and the shaded area shows the standard error. Preferences are labeled by the ground-truth reward function.}
    \label{fig:env_free_reward_real_traj_comp}
\end{figure}


\textbf{Comparison to sampling trajectories from the real environment~~} A natural concern is that EARS, by learning from imagined rather than real trajectories, might require many more preferences to learn the same reward function. We test this by comparing the data-efficiency of EARS against Offline-RLHF and Learning To Rank Real Trajectories, both of which learn from real environment trajectories, in Figure \ref{fig:env_free_reward_real_traj_comp}. We find that learning from real trajectories does not require substantially fewer preferences than learning from imagined trajectories: EARS is in fact substantially more data-efficient than Offline-RLHF, and tends to match or outperform Learning To Rank Real Trajectories. 



In Appendix~\ref{app:raw_obs} we evaluate EARS when sampling imagined trajectories from the environment observation space as opposed to the set of reward features generated in Stage 1, and find sampling imagined trajectories from the environment's observation space substantially under performs sampling imagined trajectories from the LLM designed feature space.

\subsection{Results when specifying a reward function via natural language}
\label{sec:reward_desc_rew_spec}
We now evaluate EARS in a setting where it is directly deployable. 
Here we assume the input to EARS is a natural language specification of an objective, for evaluation purposes designed from the ground-truth reward function to emulate the specifications a human stakeholder could provide. The specification is input to an LLM that will label preferences instead of a ground-truth preference labeler (Section \ref{sec:synth_pref_rew_spec}) or a human stakeholder. We evaluate EARS with the Privileged Reward Description and Realistic Reward Description outlined in Section \ref{sec:baselines}, matching the corresponding baselines. All other evaluation details match those in Section \ref{sec:synth_pref_rew_spec}.

Table \ref{tab:env_free_reward_results_llm_prefs} shows the results. EARS produces more aligned reward functions than the direct prompting baselines in nearly all settings, and yields substantially better performance when alignment with the ground-truth reward function is otherwise low. The exception is the traffic environment with the realistic reward description for $r_1$, where EARS underperforms direct prompting. Overall, EARS generally outperforms direct prompting, though at the cost of substantially more LLM generations for preference labeling.

%

\section{Conclusion}

We introduced EARS, a method for specifying reward functions from preferences without sampling trajectories from the environment---which can be prohibitively slow or unsafe. Across three sequential decision-making domains, EARS produces reward functions more aligned with the ground-truth reward function than the only existing alternatives that avoids environment sampling: directly prompting an LLM to generate a reward function. This holds in both evaluation settings we consider---when preferences are generated from a ground-truth reward function, and when they are labeled by an LLM conditioned on the kind of natural language specification a careful human stakeholder could readily provide without ever writing a reward function. EARS also matches or outperforms methods that learn from preferences labeled over real trajectories. 

Several directions for future work would extend the applicability of EARS. The setting we evaluate in Section~\ref{sec:reward_desc_rew_spec}---where a stakeholder supplies a natural language specification of the task objectives and an LLM labels preferences over imagined trajectories---requires no human in the loop and is readily practical. A natural extension is to evaluate EARS in a setting where preferences are labeled by human stakeholders rather than an LLM or a ground-truth reward function. This raises open questions our current evaluation cannot answer: whether stakeholders can meaningfully reason about imagined trajectories that may not be physically realizable. A user study with domain experts would address these questions.
While we show that LLMs can design reward functions by labeling preferences over imagined trajectories given a natural language specification, what kinds of specifications human stakeholders \emph{can} or \emph{should} provide remain important open questions. 
\bibliographystyle{plainnat}
\bibliography{main}

\newpage
\appendix
\raggedbottom

\section{Evaluating the learned reward function.}
\label{app:sec:eval_learned_rew}
To evaluate the alignment of the learned reward function with respect to a ground-truth reward function, we compute the TAC between the two as outlined in Section \ref{sec:eval_learned_rew}. An alternative method to evaluate $\hat{r}$ could be to compute the the set of optimal policies with respect to $\hat{r}$ and evaluate the average performance of those polices under the $r$: $\frac{1}{|\pi^*_{\hat{r}}|} \sum_{\pi \in \pi^*_{\hat{r}}} J_r(\pi)$. We avoid this evaluation methodology because, for the environments we consider, training any policy is computationally expensive and, due to the long task horizons and continuous state spaces, deriving an optimal policy is intractable. For a sufficiently large and diverse set of evaluation trajectories, $\mathrm{TAC}(\hat{r}, r)$ provides a strong measure of alignment; unlike empirical policy performance, it evaluates how the reward functions rank all trajectories, rather than focusing only on the observed highest-ranked ones. Broadly, the focus of our work is on producing aligned reward functions, which $\mathrm{TAC}$ evaluates directly, rather than reward functions that can train policies.

\section{Labeling synthetic preferences over imagined trajectories}
\label{app:imagined_pref_mapping}

We evaluate EARS in Section \ref{sec:synth_pref_rew_spec} by assuming access to a synthetic preference labeler that labels preferences according to the ground-truth reward function. These preferences are labeled over imagined trajectories rather than real trajectories sample from an environment, and the ground-truth reward function is defined over transitions sampled from the real environment (e.g., $r(s,a,s')$). Therefore, to enable learning from synthetic preferences over imagined trajectories, we find $r_m$ that best satisfies the ranking induced by $r$ over all trajectories in $\mathcal{T}_{\text{eval}}$. Following the Bradley--Terry formulation, we model pairwise preferences as
\[
P(\Phi_\tau \succ \Phi_{\tau'} \mid r_m) = \sigma\!\big(r_m(\Phi_\tau) - r_m(\Phi_{\tau'})\big),
\]
and learn $r_m$ by minimizing the cross-entropy loss against the ranking induced by $r$:
\begin{equation}
\label{eq:loss_rm}
\begin{aligned}
\mathcal{L}(r_m; \mathcal{T}_{\text{eval}}) = - \hspace{-2.5mm}\sum_{\tau, \tau' \in \mathcal{T}_{\text{eval}}} \hspace{-2mm}\big[ & (1 - \mu_{\tau, \tau'}) \log P(\Phi_\tau \succ \Phi_{\tau'} \mid r_m) \\
& + \mu_{\tau, \tau'} \log P(\Phi_\tau \prec \Phi_{\tau'} \mid r_m) \big],
\end{aligned}
\end{equation}
where $\mu_{\tau, \tau'} = \mathds{1}[r(\tau') > r(\tau)]$ encodes the ground-truth, noiseless preference under $r$.

To avoid arbitrary scaled of the mapped ground-truth reward function, we then scale the mapped ground-truth reward function such that the mean preference probability under the Bradley-Terry model for preferences over all trajectories in $\mathcal{T}_{\text{eval}}$ is $0.9$. Synthetic preferences over imagined trajectories can then be labeled with the mapped ground-truth reward function.

\section{Comparison to Sampling Imagined Trajectories from the Environment Observation Space}
\label{app:raw_obs}
Stage 1 of EARS aims to produce a small set of reward features from which we can sample imagined trajectories in Stage 2. Here, we compare EARS to a method that uses the environment's observation space variables as features to sample from rather than executing Stage 1. Results are shown in Table \ref{tab:alignment_raw_obs}, indicating that sampling imagined trajectories from the environment's observation space substantially under performs sampling imagined trajectories from the LLM designed feature space. The raw observation space is substantially larger, containing many redundant features unrelated to any objective. This makes informative sampling in Stage~2 harder: analogous to the noisy-TV problem in exploration, the uncertainty criterion is drawn toward pairs differing along these irrelevant features, where reward-function disagreement is high but uninformative about the ground-truth objective.

\begin{table}[tb]
\centering
\caption{\textbf{Effect/Ablation of designed reward features:~~} We compare EARS (2k stochastic prefs.), which samples imagined trajectories from the LLM-designed feature space produced in Stage 1 (denoted as Designed feats), against an ablation that instead samples imagined trajectories directly from the environment's observation space variables, bypassing Stage 1 (denoted as Raw obs). See Table \ref{tab:env_free_reward_results} for more details.}
\label{tab:alignment_raw_obs}
\setlength{\tabcolsep}{10pt}
\renewcommand{\arraystretch}{0.95}
\small
\begin{tabular}{lcccc}
\toprule
& \multicolumn{2}{c}{Pandemic} & \multicolumn{2}{c}{Glucose} \\
\cmidrule(lr){2-3} \cmidrule(lr){4-5}
Variant & $r_1$ & $r_2$ & $r_1$ & $r_2$ \\
\midrule
Designed feats & $\boldsymbol{.73\!\pm\!.01}$ & $\boldsymbol{.94\!\pm\!.02}$ & $\boldsymbol{.74\!\pm\!.03}$ & $\boldsymbol{.71\!\pm\!.06}$ \\
Raw obs & $-.42\!\pm\!.00$ & $-.05\!\pm\!.02$ & $.44\!\pm\!.01$ & $.07\!\pm\!.04$ \\
\bottomrule
\end{tabular}
\end{table}

\section{Implementation Details for LTRRT and Offline-RLHF}
\label{app:offline-pref-learning}

We compare \textsc{EARS} against two preference-learning baselines that operate
directly on environment observations rather than on LLM-generated reward
features: \emph{Learning To Rank Real Trajectories} (LTRRT) and
\emph{Offline-RLHF}. Both baselines are trained and evaluated entirely offline
from a fixed pool of pre-collected trajectories. The two baselines are identical in every respect---trajectory pool,
preference labeling, reward-model architecture, optimizer, and
evaluation---and differ \emph{only} in how a trajectory is encoded into the
fixed-length feature vector consumed by the reward model. This isolates the
effect of the trajectory representation.

\subsection{Preference-pair construction}
For each environment we use a fixed dataset of $450$ pre-collected trajectories, which are sampled from various checkpoints when training a policy with the ground-truth reward functions. Each trajectory is a sequence of
$(s,a,s')$ transitions. We uniformly sample pairs of trajectories from the fixed dataset of $900$ trajectories.

\subsection{Trajectory representations (the only difference between baselines)}
Let $\phi(s_t, a_t, s_{t+1}) \in \mathbb{R}^{d}$ denote the reward feature vector for a single transition $(s_t, a_t, s_{t+1})$.

\paragraph{LTRRT (Learning To Rank Real Trajectories).}
A trajectory is summed over its transitions into a single $d$-dimensional vector,
$\Phi^{\text{LTRRT}} = \sum_{t=0}^{|\tau|-1} \phi(s_t, a_t, s_{t+1}) \in \mathbb{R}^{d}$, and then min--max
normalized per dimension using precomputed per-feature ranges $[\ell_m, h_m]$,
i.e. $\Phi^{\text{LTRRT}}_m \leftarrow (\Phi^{\text{LTRRT}}_m - \ell_m)/(h_m-\ell_m)$.
The model input dimension is therefore $d$, independent of trajectory length.

\paragraph{Offline-RLHF.}
A trajectory is encoded by concatenating the per-transition feature vectors across
all transitions,
$\Phi^{\text{RLHF}} = [\phi(s_0, a_0, s_1); \phi(s_1, a_1, s_2); \dots; \phi(s_{|\tau|-1}, a_{|\tau|-1}, s_{|\tau|})] \in \mathbb{R}^{|\tau|\cdot d}$.  These features are \emph{not} normalized.
This representation preserves the full per-transition feature trace, at the
cost of a much higher input dimension.

\subsection{Training details}
Both baselines learn a model $f_\theta(\cdot)$
implemented as a neural network. The network has $2$ hidden layers of width $64$ with $\tanh$ activations, followed by
a linear layer to a single scalar output. The input dimension equals the representation dimension of the corresponding
baseline. The two baselines
use exactly the same architecture and hyperparameters; only the input dimension differs.

The reward model is trained by minimizing the pairwise binary cross-entropy loss from Eq. \ref{eq:loss}. We use Adam with learning rate $10^{-3}$, weight decay $0$, for $1000$ epochs.

\section{EARS Hyperparameters}
\label{app:ears-hyperparameters}

This section details the hyperparameters used for \textsc{EARS} in both stages
and, where relevant, how they were chosen. Because \textsc{EARS} never samples
the environment, it exposes very few hyperparameters that require tuning:
the reward-feature design in Stage~1 is driven by an LLM conversation with 
sampling temperature $1$, and the weight-learning procedure
in Stage~2 inherits the defaults from the preference-learning and epistemic
neural network (ENN) literature. We tuned \emph{no} hyperparameter against the
evaluation metric (TAC); all values below are either standard library defaults,
choices dictated by the problem structure, or design decisions fixed a priori
for reasons we state explicitly. The same settings are used across all three
environments and both ground-truth reward functions unless noted.

\subsection{Stage 1: Designing reward features}
\label{app:ears-stage1}

\paragraph{LLMs and generation settings.}
Stage~1 is a structured conversation between a stakeholder LLM and a
facilitator LLM. Both roles are served by the same model,
\texttt{gemini-3-pro-preview}, accessed through Vertex AI. We do not
override any decoding parameters (temperature, top-$p$, or maximum output
tokens); all generations use the provider defaults. We deliberately avoid
lowering the temperature so that the deliberation explores a diverse set of
candidate objectives rather than collapsing onto a single mode.

\paragraph{Conversation structure and termination.}
The stakeholder--facilitator dialogue is not an open-ended chat with a
``maximum number of turns'' hyperparameter; it is a fixed finite-state graph of
prompt nodes that (i) elicit candidate objectives, (ii) filter them to those
measurable from the observation space, and (iii) implement each surviving
objective as an executable Python reward feature. 

\paragraph{Number of reward features $d$.}
The number of reward features $d$ is \emph{not} a hyperparameter. We never
specify a target or maximum count; the stakeholder is prompted to produce a
``small set'' of measurable objectives, each implemented as a separate feature
(with one-hot expansion for categorical objectives), and $d$ emerges from the
deliberation. This keeps the feature set interpretable without imposing an
arbitrary dimensionality.

\paragraph{Feature ranges $\phi^H_{\min},\phi^H_{\max}$.}
Stage~1 also generates, for each feature, a speculative minimum and maximum of its
sum over a horizon of $H$ steps, which defines the range that Stage~2 samples
imagined trajectories from. These are produced in two steps. First, the
stakeholder specifies a per-transition output range for each feature (a discrete set
such as $\{-1,0,1\}$, a bounded interval such as $[0,1]$, or a semi-/unbounded
interval). Second, a separate LLM call converts each per-transition range into a
finite episode-level range by applying the convention that a feature summed over
$H$ steps with per-step range $[a,b]$ has episode range $[Ha,Hb]$, with
an exception for features that do not accumulate additively (e.g., a
binary failure indicator remains $\{0,1\}$). The horizons are fixed by the
environments and are not tunable: $H=192$ (Pandemic), $H=5760$ (Glucose), and
$H=300$ (Traffic).

\subsection{Stage 2: Learning weights from preferences}
\label{app:ears-stage2}

\paragraph{Sampling imagined trajectories.}
Each imagined trajectory $\Phi\in\mathbb{R}^d$ is drawn by sampling each feature
independently and uniformly over its Stage-1 episode range
$[\phi^H_{\min},\phi^H_{\max}]$: continuous features via a uniform draw, discrete
features via a uniform choice over the allowed values, and categorical groups so
that the group sums to its ceiling. Candidate pairs are restricted to differ in
\emph{at most two features}. This restriction is fixed a priori---not tuned---on
the hypothesis that a labeler can reason more reliably about a pair whose salient
differences are confined to one or two features; it is enforced structurally by
zeroing all but the one or two active feature coordinates when constructing a
candidate pair.

\paragraph{Learning weights from noiseless preferences (linear program).}
When preferences are assumed noiseless, we learn $\hat{w}$ by solving the
feasibility linear program of Eq.~4 with \texttt{scipy.optimize.linprog}
(HiGHS backend). The tolerance that turns each non-strict preference inequality
into a strict one is $\epsilon=10^{-4}$. Each weight is box-constrained to
$w_i\in[-1,1]$ (playing the role of the norm bound $B$ in Eq.~4), and the
resulting solution is $\ell_1$-normalized before it is used as a reward
function. The acquisition score is the LP-disagreement measure $f_{\mathrm{LP}}$
of Section~3.2: a pair is considered informative only when the
minimum and maximum signed return differences over the feasible weight set have
opposite signs. $\epsilon$ is set small enough to enforce strictness without
distorting the feasible set.

\paragraph{Learning weights from stochastic (Bradley--Terry) preferences.}
When preferences follow the Bradley--Terry model, uncertainty is quantified as
the variance in predicted preference probability across an ENN ensemble of
reward models~(Osband et al.\ 2023). The ENN is an ensemble MLP with
$M=10$ members, two hidden layers of width $64$ with ReLU activations, and a
scalar output; the epistemic index selects an ensemble member (index dimension
$=M$). It is trained with the Bradley--Terry loss
averaged over $10$ epistemic-index samples, using Adam with learning rate
$10^{-2}$, an $\ell_2$ loss-side penalty of $10^{-3}$, and batch size $16$;
after each new preference the ensemble is refit for $10$ gradient steps over all
accumulated pairs. At acquisition time the preference-probability variance is
estimated from the $M=10$ ensemble members (with the prior scale set to $1.0$).
The final reward weights $\hat{w}$ reported for the stochastic setting are then
obtained by fitting a single \emph{linear} model to the collected preferences
with the binary cross-entropy loss of Eq.~2, optimized with L-BFGS
(strong-Wolfe line search, up to $1{,}000$ iterations) with no additional
regularization and $5$ random restarts keeping the lowest-loss solution. These
ENN settings are the library defaults for preference-based ENNs and were not
tuned to our environments.

\paragraph{Preference labeling.}
For the synthetic-preference experiments (Section~4.4), preference labels come
from the ground-truth reward function; to make label noise comparable across
environments and reward functions with very different return scales, the mapped
ground-truth reward is calibrated so that the mean Bradley--Terry preference
probability over the evaluation trajectories equals a target of $0.9$
(Appendix~B). For the natural-language experiments (Section~4.5), preferences are
labeled by \texttt{gemini-3-pro-preview} conditioned on the specification; each
pair is labeled by majority vote over $3$ independent LLM queries.

\subsection{Compute and software environment}
\label{app:ears-hardware}

All \textsc{EARS} experiments were run on two nodes of an internal compute
cluster. Each node runs Ubuntu~24.04.4~LTS with two
Intel Xeon E5-2680~v4 CPUs at $2.40$\,GHz ($28$ physical cores / $56$ hardware
threads in total), roughly $250$\,GB of system RAM, and an NVIDIA GeForce
GTX~1080~Ti GPU ($11$\,GB). Consistent with our problem setting, \textsc{EARS}
is inexpensive to run locally: because it never samples the environment, its only
local computation is the lightweight reward-weight learning of Stage~2---feasibility
linear-program solves and small ENN ensemble fits---which is CPU-bound and does
not require the GPU. The dominant computational cost is LLM inference.

\section{Ground-truth reward function descriptions}
\label{app:gt_rew_desc}

The ground-truth reward functions we use as alignment targets are given below for all environments: 

\eqboxlabel{fig:traffic_r1}
\begin{eqbox}
\begin{equation*}
r_1 \;=\;
\begin{cases}
0, & \text{if collision} \\[2pt]
\eta_1\, c_1 \;+\; \eta_2\, c_2 \;+\; \eta_3\, c_3, & \text{otherwise}
\end{cases}
\end{equation*}
\begin{align*}
c_1 &= \frac{\max\!\big(v^\star\sqrt{N}\,-\,\lVert \mathbf{v}-v^\star\mathbf{1}\rVert_2,\ 0\big)}{v^\star\sqrt{N}}, \\
c_2 &= \sum_{i:\,v_i^{\mathrm{ego}}>0}\,
        \min\!\Bigg(\frac{h_i / v_i^{\mathrm{ego}} \,-\, t_{\min}}{t_{\min}},\ 0\Bigg), \\
c_3 &= \min\!\Bigg(\alpha \;-\; \tfrac{1}{M}\sum_{j=1}^{M} |a_j|,\ 0\Bigg).
\end{align*}
\tcblower
\small
\textbf{Traffic Control Ground-truth reward function $r_1$.} $\mathbf{v}\in\mathbb{R}^N$ are the speeds of all $N$ vehicles; $v^\star$ is the target velocity; $v_i^{\mathrm{ego}}$ and $h_i$ are the speed and leader headway of the $i$-th RL vehicle; $\mathbf{a}\in\mathbb{R}^M$ are the accelerations of the $M$ RL vehicles; $t_{\min}=1$\,s; $\alpha=0.1$; $(\eta_1,\eta_2,\eta_3)=(1.0,\,0.1,\,1.0)$.
\end{eqbox}

\bigskip

\eqboxlabel{fig:traffic_r2}
\begin{eqbox}
\begin{equation*}
r_2 \;=\; \frac{1}{N}\sum_{i=1}^{N} v_i .
\end{equation*}
\tcblower
\small
\textbf{Traffic Control Ground-truth reward function $r_2$.} $v_i$ is the speed of the $i$-th vehicle and $N$ is the total number of vehicles in the simulator.
\end{eqbox}

\bigskip


\eqboxlabel{fig:pandemic_r1}
\begin{eqbox}
\begin{equation*}
\begin{aligned}
r_1 \;=\;\;
& -10\, C_t
\;-\; 10\,\mathds{1}[\,s_t - s_{t-1}=1\,]
\bigg(\frac{\min(\rho_{t-1}-0.005,\,0)}{0.005}\bigg)^{\!2} \\
&\;-\; 0.1\,\frac{s_t^{1.5}}{4^{1.5}}
\;-\; 0.02\,|s_t - s_{t-1}| .
\end{aligned}
\end{equation*}
\tcblower
\small
\textbf{Pandemic Mitigation Ground-truth reward function $r_1$.} $s_t\in\{0,1,2,3,4\}$ is the regulation stage at step $t$; $C_t$ is the population-mean fraction of critical cases at step $t$; $\rho_{t-1}=I_{t-1}+C_{t-1}+D_{t-1}$ is the sum of the infected, critical, and dead population fractions at the previous step; $\mathds{1}[\cdot]$ is the indicator function.
\end{eqbox}

\bigskip

\eqboxlabel{fig:pandemic_r2}
\begin{eqbox}
\begin{equation*}
r_2 \;=\; -10\, C_t \;-\; 0.1\,\frac{s_t^{1.5}}{4^{1.5}} \;-\; 0.02\,|s_t - s_{t-1}| .
\end{equation*}
\tcblower
\small
\textbf{Pandemic Mitigation Ground-truth reward function $r_2$.} $s_t\in\{0,1,2,3,4\}$ is the regulation stage at step $t$; $C_t$ is the population-mean fraction of critical cases at step $t$.
\end{eqbox}

\bigskip


\eqboxlabel{fig:glucose_r1}
\begin{eqbox}
\begin{equation*}
r_1 \;=\;
\begin{cases}
-10^{13}, & \text{if terminated} \\[3pt]
-10\,\Big[\,3.5506\big(\ln(\max(b_t,1))^{0.8353} -\,3.7932\big)\Big]^{2}\;-\; 10\, u_t, & \text{otherwise}
\end{cases}
\end{equation*}
\tcblower
\small
\textbf{Glucose Monitoring Ground-truth reward function $r_1$.} $b_t$ is the most recent blood-glucose reading (mg/dL); $u_t$ is the most recent insulin dose; the terminated case corresponds to patient death.
\end{eqbox}

\bigskip

\eqboxlabel{fig:glucose_r2}
\begin{eqbox}
\begin{equation*}
r_2 \;=\;
\begin{cases}
-10^{13}, & \text{if terminated} \\[3pt]
-0.32\, u_t \;-\; \mathds{1}[\,b_t < 70\,]\,\dfrac{10\cdot 1350}{12\cdot 24\cdot 365}, & \text{otherwise}
\end{cases}
\end{equation*}
\tcblower
\small
\textbf{Glucose Monitoring Ground-truth reward function $r_2$.} $b_t$ is the most recent blood-glucose reading (mg/dL); $u_t$ is the most recent insulin dose; $\mathds{1}[b_t<70]$ flags hypoglycemia; the terminated case corresponds to patient death.
\end{eqbox}

As an input to the Direct Prompting baseline, and to the LLM-preference labeler that is used for EARS in Section \ref{sec:reward_desc_rew_spec}, we use a natural language description of the ground-truth reward function. The Privileged Reward Description, which we hypothesize is likely more specific than a human stakeholder could reasonably provide, is given in Figures \ref{fig:pandemic_r1_pt1} and \ref{fig:pandemic_r2_pt1} for Pandemic Mitigation, Figures \ref{fig:glucose_r1_pt1} and \ref{fig:glucose_r2_pt1} for Glucose Monitoring, and Figures \ref{fig:traffic_r1_pt1} and \ref{fig:traffic_r2_pt1} for Traffic Control. 

The Realistic Reward Description, which we hypothesize is a specification a human stakeholder could reasonably provide, is given in Figures \ref{fig:pandemic_r1_pt2} and \ref{fig:pandemic_r2_pt2} for Pandemic Mitigation, Figures \ref{fig:glucose_r1_pt2} and \ref{fig:glucose_r2_pt2} for Glucose Monitoring, and Figures \ref{fig:traffic_r1_pt2} and \ref{fig:traffic_r2_pt2} for Traffic Control.


\begin{figure}[H]
\centering
\begin{summarybox}
The reward function should aggregate four distinct penalty components to guide pandemic management policies effectively. First, it should significantly penalize the proportion of the population currently in a critical infection state to prioritize public health. Second, it must apply a specific political penalty for increasing regulation stringency when the infection rate is negligible or very low. Third, it should discourage the maintenance of strict regulation stages, imposing a penalty that scales with the severity of the current restrictions. Finally, the function should penalize the magnitude of changes in regulations between time steps to encourage smooth and stable policy transitions.
 
The tradeoffs between the features of the reward function are as follows:
\begin{itemize}\itemsep0pt
    \item Increasing the proportion of critical cases in the population by 0.01 units is equivalent to increasing the normalized stage severity penalty by 1.0 unit.
    \item Increasing the magnitude of the regulation stage change by 1.0 unit is equivalent to increasing the normalized stage severity penalty by 0.2 units.
    \item Increasing the calculated political penalty factor by 1.0 unit is equivalent to increasing the proportion of critical cases in the population by 1.0 unit.
\end{itemize}
\end{summarybox}
\caption{Pandemic Mitigation Privileged Reward Description for ground-truth reward function $r_1$}
\label{fig:pandemic_r1_pt1}
\end{figure}
 
\begin{figure}[H]
\centering
\begin{summarybox}
\textbf{What outcomes should the agent try to achieve?}
\begin{itemize}\itemsep0pt
    \item \textbf{Minimize severe health consequences:} The primary goal is to keep the number of critical infection cases as low as possible. We want to prevent the healthcare system from being overwhelmed.
    \item \textbf{Maintain normalcy/freedom:} The agent should aim to keep the restriction level as low as possible. If it is safe to do so, the preference is always for looser regulations rather than strict lockdowns.
    \item \textbf{Policy consistency:} The agent should try to keep the chosen restriction level stable over time. People need to be able to plan their lives, so we want to see long stretches of the same stage rather than constant changes.
\end{itemize}
 
\textbf{What outcomes should the agent avoid?}
\begin{itemize}\itemsep0pt
    \item \textbf{High numbers of critical patients:} Avoid situations where a large portion of the population falls critically ill.
    \item \textbf{Severe lockdowns:} Avoid using the highest levels of government intervention unless it is the only option left.
    \item \textbf{Erratic ``flip-flopping'':} Avoid changing the restriction stage abruptly or frequently. For example, don't jump from no restrictions to a full lockdown and back again within a short period.
    \item \textbf{Unjustified tightening:} Specifically avoid increasing the restriction level when the current number of infections is actually low. It looks bad politically and socially to impose harsher rules when the virus isn't actively surging.
\end{itemize}
 
\textbf{Describe how the agent should trade off between different objectives or outcomes.}
\begin{itemize}\itemsep0pt
    \item While public health is critical, the agent needs to carefully balance it against the economic and social cost of lockdowns. Do not choose a high restriction level just to prevent a very small number of cases; only use strict measures if the critical cases are becoming significant.
    \item Prioritize consistency over immediate reaction. Even if a slight change might be optimal for infection numbers, it is often better to stay the course to avoid confusing the public.
    \item Be very hesitant to increase restrictions if the infection rate is currently low. The ``cost'' of adding restrictions feels much higher to the population when they don't see an immediate threat, so the agent should really resist doing that.
\end{itemize}
\end{summarybox}
\caption{Pandemic Mitigation Realistic Reward Description for ground-truth reward function $r_1$}
\label{fig:pandemic_r1_pt2}
\end{figure}
 
\begin{figure}[H]
\centering
\begin{summarybox}
The reward function should guide the agent to balance public health outcomes with the economic and social costs of restrictions and policy instability. It must penalize the presence of critical infection cases within the population to prioritize health safety. Simultaneously, the function should penalize the severity of the current regulation stage, encouraging the agent to keep restrictions as low as possible. Additionally, it must discourage frequent or drastic changes to regulations by penalizing the difference in stages between time steps to ensure smooth policy transitions.
 
The tradeoffs between these features interact as follows: increasing the proportion of the population in a critical condition by 1\% (0.01) is equivalent to the penalty of imposing the maximum level of regulation severity. Furthermore, increasing the critical population proportion by just 0.2\% (0.002) results in a penalty equivalent to changing the regulation stage by a single level. Consequently, the penalty for maintaining the maximum regulation severity is equivalent to the penalty accumulated by changing the regulation stage 5 times.
\end{summarybox}
\caption{Pandemic Mitigation Privileged Reward Description for ground-truth reward function $r_2$}
\label{fig:pandemic_r2_pt1}
\end{figure}
 
\begin{figure}[H]
\centering
\begin{summarybox}
\textbf{What outcomes should the agent try to achieve?}
\begin{itemize}\itemsep0pt
    \item Keep the number of people who are critically ill as low as possible. The primary goal is to save lives and prevent the healthcare system from getting overwhelmed, so focus on minimizing the number of critical infection cases.
    \item Keep society as open as possible. Whenever it is safe to do so, the agent should aim for the lowest level of restrictions (like Stage 0 or Stage 1) to minimize the impact on the economy and people's daily lives.
\end{itemize}
 
\textbf{What outcomes should the agent avoid?}
\begin{itemize}\itemsep0pt
    \item Avoid situations where the number of severe infections spikes uncontrollably. We don't want to reach a point where hospitals cannot cope with the influx of patients.
    \item Avoid erratic or ``flip-flopping'' policies. The agent should avoid drastically changing the restriction level back and forth over short periods (e.g., going from Level 1 to Level 4 and back to Level 1 in a few days), as this causes confusion and instability.
\end{itemize}
 
\textbf{Describe how the agent should trade off between different objectives or outcomes.}
\begin{itemize}\itemsep0pt
    \item Public health is the priority, but not at any cost. If critical cases are rising and threatening the healthcare system, the agent must accept higher restrictions to bring numbers down. However, if the health situation is stable or critical cases are low, the agent should prioritize lowering restrictions to relieve the social and economic burden.
    \item Regarding policy changes: The agent should favor stability. It is better to maintain a slightly higher restriction level for a bit longer than to lower it and immediately have to raise it again. Only make large jumps in restrictions if there is an extreme emergency; otherwise, prefer steady, predictable changes.
\end{itemize}
\end{summarybox}
\caption{Pandemic Mitigation Realistic Reward Description for ground-truth reward function $r_2$}
\label{fig:pandemic_r2_pt2}
\end{figure}


\begin{figure}[H]
\centering
\begin{summarybox}
The reward function should serve to maintain healthy blood glucose levels while minimizing insulin usage, with a strict priority on preventing patient death. If the simulation terminates, a very large negative penalty must be applied immediately. Otherwise, the function should calculate a risk metric based on the current blood glucose level using a non-linear transformation that penalizes deviations from the target range. This risk score is then combined with a penalty proportional to the amount of insulin administered to form a negative cost function. The final reward is the negative sum of the glucose risk and the insulin penalty. In terms of tradeoffs, increasing the squared transformed glucose risk value by one unit is equivalent to increasing the insulin dosage by one unit.
\end{summarybox}
\caption{Glucose Monitoring Privileged Reward Description for ground-truth reward function $r_1$}
\label{fig:glucose_r1_pt1}
\end{figure}
 
\begin{figure}[H]
\centering
\begin{summarybox}
\textbf{What outcomes should the agent try to achieve?}
\begin{itemize}\itemsep0pt
    \item Keep the patient's blood glucose levels within a healthy target range. The primary goal is to maintain stable blood sugar levels that mimic a healthy person's pancreas.
    \item Administer insulin efficiently. The agent should try to achieve the target glucose levels using the least amount of medication necessary to get the job done.
\end{itemize}
 
\textbf{What outcomes should the agent avoid?}
\begin{itemize}\itemsep0pt
    \item The agent must absolutely avoid any state that leads to the termination of the simulation or the death of the patient. This is the worst-case scenario.
    \item Avoid letting blood glucose levels get dangerously high (hyperglycemia) or dangerously low (hypoglycemia). The further the blood sugar drifts from the ideal target, the worse it is for the patient.
    \item Avoid ``insulin stacking'' or using excessive amounts of the drug when it isn't strictly needed, as this increases the risk of a crash later.
\end{itemize}
 
\textbf{Describe how the agent should trade off between different objectives or outcomes.}
\begin{itemize}\itemsep0pt
    \item The most critical priority is preventing the simulation from ending in failure or patient death; this outweighs everything else.
    \item When balancing blood sugar management against medication usage, prioritize keeping the blood sugar in the safe range. However, if two strategies result in similar blood glucose levels, prefer the one that uses less insulin to minimize the burden on the patient.
    \item Being slightly outside the target range is bad, but being extremely far outside the target range (very high or very low) is significantly worse and should be penalized much more heavily.
\end{itemize}
\end{summarybox}
\caption{Glucose Monitoring Realistic Reward Description for ground-truth reward function $r_1$}
\label{fig:glucose_r1_pt2}
\end{figure}
 
\begin{figure}[H]
\centering
\begin{summarybox}
The reward function should serve as an economic optimizer that treats the agent's actions and physiological state as financial costs to be minimized. It must assign a catastrophic negative value if the patient does not survive, ensuring that maintaining the patient's life is the paramount constraint. Routine usage of insulin should be penalized proportionally to the amount administered, reflecting the direct financial cost of the medication. Additionally, the function should apply a discrete penalty whenever blood glucose levels drop below a critical safety threshold, simulating the probabilistic cost of emergency medical care or hospitalization. The overall objective is to maximize the negative sum of these expenses, encouraging the agent to balance medication costs against the financial risks of severe health complications.
 
In terms of feature tradeoffs, the reward is composed of negative costs associated with insulin usage and hypoglycemic events. Specifically, the financial penalty for administering 1 unit of insulin is approximately 2.5 times greater than the penalty incurred for a single time-step of hypoglycemia. Therefore, avoiding a single instance of low blood glucose is equivalent to saving approximately 0.4 units of insulin.
\end{summarybox}
\caption{Glucose Monitoring Privileged Reward Description for ground-truth reward function $r_2$}
\label{fig:glucose_r2_pt1}
\end{figure}
 
\begin{figure}[H]
\centering
\begin{summarybox}
\textbf{What outcomes should the agent try to achieve?}
\begin{itemize}\itemsep0pt
    \item The primary goal is to minimize the overall financial burden of treating the patient. I want the agent to find a strategy that results in the lowest possible total cost over the treatment period.
    \item The agent should manage the patient's condition efficiently, using resources only when necessary to keep costs down.
\end{itemize}
 
\textbf{What outcomes should the agent avoid?}
\begin{itemize}\itemsep0pt
    \item Most importantly, the agent must strictly avoid any situation where the patient dies or reaches a fatal state. This is unacceptable.
    \item Avoid letting the patient's blood sugar drop too low (hypoglycemia). This is crucial because low blood sugar usually requires expensive emergency interventions or hospital visits.
    \item Avoid wasting insulin. Every unit of insulin administered costs money, so the agent shouldn't administer it unless the benefit outweighs the direct cost of the drug.
\end{itemize}
 
\textbf{Describe how the agent should trade off between different objectives or outcomes.}
\begin{itemize}\itemsep0pt
    \item Survival is the hard constraint; never let the patient reach a terminal state regardless of cost.
    \item Beyond survival, treat medical decisions as financial calculations. The agent should balance the direct cost of giving insulin against the potential future cost of emergency care.
    \item It is acceptable to allow higher blood sugar levels if bringing them down would cost more in medication than it saves in avoided hospital fees, provided the patient stays alive. The strategy should essentially be: ``Spend the least amount of money possible while keeping the patient alive.''
\end{itemize}
\end{summarybox}
\caption{Glucose Monitoring Realistic Reward Description for ground-truth reward function $r_2$}
\label{fig:glucose_r2_pt2}
\end{figure}


\begin{figure}[H]
\centering
\begin{summarybox}
The reward function should encourage efficient, safe, and smooth traffic flow by combining a system-wide velocity objective with specific penalties for unsafe or aggressive driving behaviors. It must calculate a positive, normalized score based on how closely the speeds of all vehicles in the network align with a desired target velocity. Simultaneously, the function should enforce safety by applying a penalty whenever an RL agent's time-headway (distance normalized by speed) relative to its leader falls below a minimum safe threshold. To ensure passenger comfort and fuel efficiency, it must also deduct a penalty if the average absolute acceleration of the agents exceeds a defined comfort limit. In the event of a collision or system failure, the function should immediately disregard these components and return a reward of zero.
 
The features of the reward function interact with specific tradeoffs between system efficiency, safety spacing, and driving smoothness. The global velocity score and the acceleration penalty carry equal weight (1:1); therefore, increasing the normalized velocity score by 1 unit is equivalent to reducing the excess acceleration violation (above the comfort threshold) by 1 unit. However, the penalty for unsafe time-headway is weighted significantly lower (at 10\% of the other terms), meaning that increasing the velocity score by just 0.1 units is equivalent to improving the normalized time-headway violation metric by a full 1.0 unit.
\end{summarybox}
\caption{Traffic Control Privileged Reward Description for ground-truth reward function $r_1$}
\label{fig:traffic_r1_pt1}
\end{figure}
 
\begin{figure}[H]
\centering
\begin{summarybox}
\textbf{What outcomes should the agent try to achieve?}
\begin{itemize}\itemsep0pt
    \item \textbf{Efficient Traffic Flow:} The primary goal is to keep the entire fleet moving well. I want the vehicles to try to match the target speed of the highway so that traffic doesn't bottle up. It's important that the whole system flows nicely, not just individual cars speeding up.
    \item \textbf{Safe Following Distances:} Vehicles should maintain a safe gap behind the car in front of them. They need to leave enough room (time gap) to react safely if the lead car slows down.
    \item \textbf{Passenger Comfort:} The vehicles should drive smoothly. This means accelerating and decelerating in a way that feels natural and isn't jarring for passengers.
\end{itemize}
 
\textbf{What outcomes should the agent avoid?}
\begin{itemize}\itemsep0pt
    \item \textbf{Collisions and Failures:} The most obvious one is that the cars must not crash into each other or go off the road. This is a critical failure.
    \item \textbf{Tailgating:} Avoid getting too close to the vehicle ahead. Following too closely is unsafe and stressful.
    \item \textbf{Jerky Driving:} Avoid sudden, erratic movements. The agents shouldn't be slamming on the brakes or flooring the gas pedal unless absolutely necessary to avoid a crash.
\end{itemize}
 
\textbf{Describe how the agent should trade off between different objectives or outcomes.}
\begin{itemize}\itemsep0pt
    \item Safety is the absolute baseline. If a collision or a major failure happens, the attempt counts for nothing---it should receive no reward at all (or a zero score).
    \item As long as no accidents occur, you should balance the other goals. I want the cars to be efficient and reach the target speed, but not if it means driving aggressively (jerky) or following unsafely (tailgating). Try to find a sweet spot where traffic moves fast but the ride remains smooth and the gaps between cars stay safe.
\end{itemize}
\end{summarybox}
\caption{Traffic Control Realistic Reward Description for ground-truth reward function $r_1$}
\label{fig:traffic_r1_pt2}
\end{figure}
 
\begin{figure}[H]
\centering
\begin{summarybox}
The reward function should prioritize maximizing the overall traffic flow efficiency across the entire network. To achieve this, it should calculate a reward based on the speeds of all vehicles present in the environment, encompassing both the reinforcement learning agents and the uncontrolled background traffic. The specific signal provided to the agent should be the average velocity of this collective fleet. This design encourages the agents to cooperate in a way that prevents congestion and maintains high speeds for every driver on the road. Since the reward function consists solely of the global average speed, there are no tradeoffs between competing objectives like safety or energy consumption; consequently, increasing the speed of any single vehicle by a specific amount contributes exactly the same value to the total reward as increasing the speed of any other vehicle by that same amount.
\end{summarybox}
\caption{Traffic Control Privileged Reward Description for ground-truth reward function $r_2$}
\label{fig:traffic_r2_pt1}
\end{figure}
 
\begin{figure}[H]
\centering
\begin{summarybox}
\textbf{What outcomes should the agent try to achieve?}
\begin{itemize}\itemsep0pt
    \item Keep traffic moving as fast as possible. The main goal is to have a high driving speed for every car on the road, not just the autonomous ones.
    \item Ensure the overall flow of traffic is efficient. When the AVs merge, the cars on the main highway should be able to keep driving quickly without having to slow down significantly.
    \item Improve the throughput of the whole road network. I want to see the fleet acting in a way that helps everyone get to their destination faster.
\end{itemize}
 
\textbf{What outcomes should the agent avoid?}
\begin{itemize}\itemsep0pt
    \item Avoid causing traffic jams or bottlenecks. The AVs shouldn't do anything that causes a ``shockwave'' of braking behind them.
    \item Avoid situations where cars are stuck or crawling along. Any situation where the average speed of the cars drops to near zero is bad.
    \item Avoid collisions, obviously, but also avoid forcing human drivers on the highway to slam on their brakes to let the AVs in, as this creates congestion.
\end{itemize}
 
\textbf{Describe how the agent should trade off between different objectives or outcomes.}
\begin{itemize}\itemsep0pt
    \item The most important thing is the collective speed of all cars. It is acceptable for one specific autonomous vehicle to slow down or wait a moment on the ramp if doing so prevents a traffic jam on the main highway.
    \item Prioritize the ``big picture'' flow over the speed of any single vehicle. If slowing down one car allows ten others to maintain a high speed, the agent should do that. The goal is maximizing the speed of the group, not just the individual.
\end{itemize}
\end{summarybox}
\caption{Traffic Control Realistic Reward Description for ground-truth reward function $r_2$}
\label{fig:traffic_r2_pt2}
\end{figure}

\end{document}